\documentclass[11pt,a4paper]{article}

\usepackage[T1]{fontenc}
\usepackage[utf8]{inputenc}
\usepackage{lmodern}

\usepackage[a4paper,
            left=22mm,right=22mm,
            top=22mm,bottom=24mm,
            headheight=14pt]{geometry}

\usepackage{amsmath,amssymb,amsfonts}
\usepackage{mathtools}

\usepackage{booktabs}
\usepackage{array}
\usepackage{multirow}
\usepackage{tabularx}
\usepackage{caption}
\usepackage{graphicx}
\usepackage{float}

\usepackage[dvipsnames]{xcolor}

\usepackage{listings}
\usepackage[colorlinks=true,
            linkcolor=blue!60!black,
            citecolor=blue!60!black,
            urlcolor=blue!60!black,
            pdfborder={0 0 0}]{hyperref}

\usepackage{enumitem}
\setlist{nosep,leftmargin=*}
\usepackage{microtype}

\title{\textbf{Two Is Enough: Binary Classification Priors Suffice\\
for Multi-Class In-Context Tabular Prediction}\\[2pt]
\large An Empirical Study under the TabICLv2 Framework}
\author{Zhixiong Cheng, Shiyuan Han, Jingjing Suo\thanks{Corresponding author. Email: \href{mailto:jingjing.suo@brgroup.com}{jingjing.suo@brgroup.com}}, Zhengli Wang \\
\small Bairong Inc.AI, Beijing, China}
\date{2026-10-20}

\begin{document}
\maketitle

\begin{abstract}
\noindent
Tabular foundation models (TFMs) pretrain a transformer on large-scale synthetic priors so that a single forward pass yields zero-shot tabular prediction. TabICLv2 samples the class count of its classification prior during pretraining, and the prevailing configuration caps it at \textbf{ten}. We report a counter-intuitive result: setting this cap \textbf{directly to two}---making every pretrained classification task binary---still yields weights that perform zero-shot \textbf{K-way} classification ($K \gg 2$). On the 38 TabArena classification datasets (200 bootstrap rounds, anchored at default RF = 1000), the binary two-stage recipe reaches \textbf{1590.1 Elo}, above the official default at 1568.0 ($\boldsymbol{\Delta = +22.1}$), above TabPFN-v2.6 (1577.1) and RealTabPFN-v2.5 (1521.6). A production-scale validation on 60 real-world credit-scoring institutions independently confirms the direction: the same checkpoint improves test AUC on 44 of 60 institutions (paired Wilcoxon $p = 0.0007$ for AUC, $p = 0.0002$ for KS) at $+1.5\%$ inference cost. Mechanistically, classification in TabICLv2 is carried not by a \emph{fixed-cardinality output head} but by an \emph{in-context comparison operator} matching query-row embeddings against class-conditional embeddings, so the cardinality of the output space is set by the \textbf{labels present in the context}, not by the label space seen during pretraining. This makes a binary prior structurally cheaper to generate and cache than a ten-class prior, demotes ``number of classes'' from an \textbf{architectural constraint} to a \textbf{tunable hyperparameter}, and points to a class-count-agnostic regime of TFM pretraining.
\end{abstract}

% =============================================================================
\begin{figure}[H]
\centering
\includegraphics[width=0.8\textwidth]{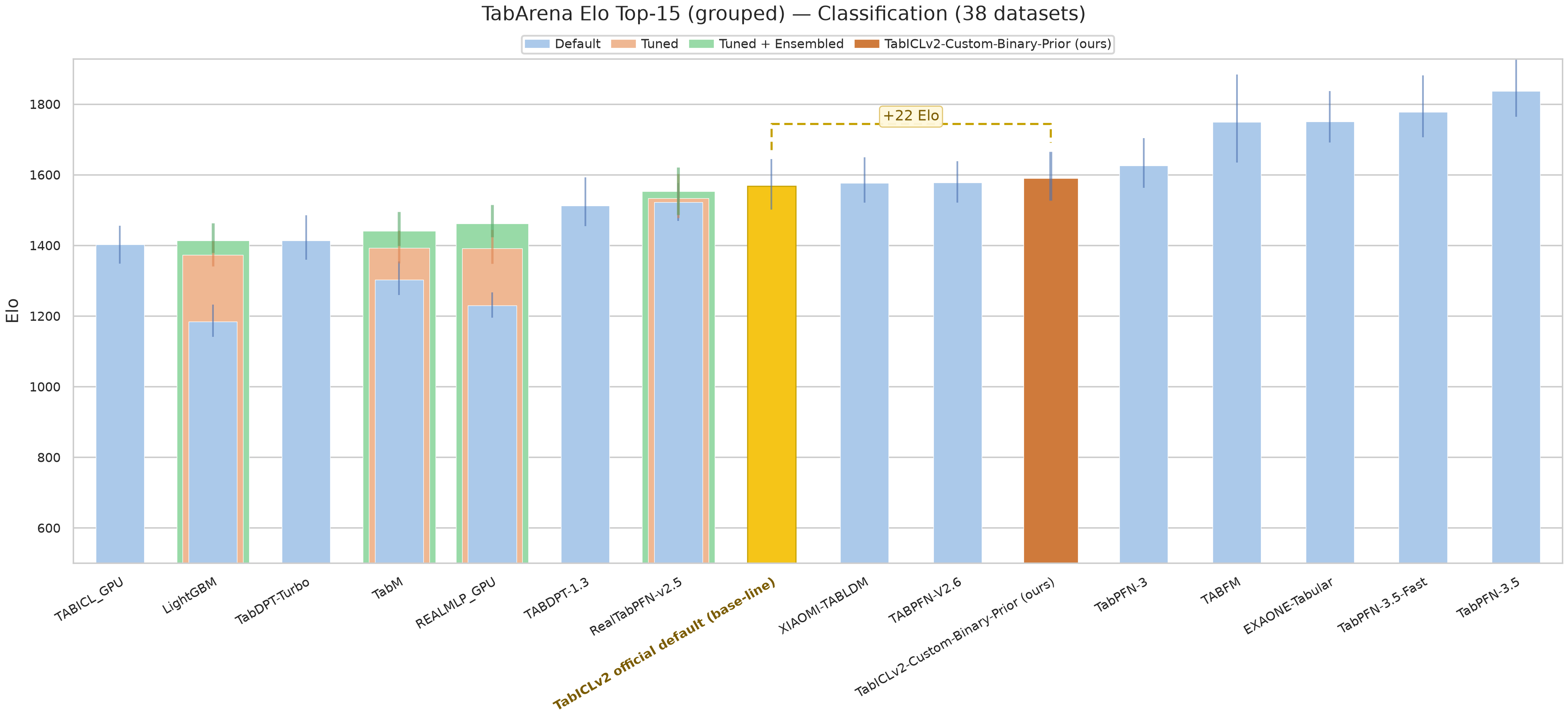}
\caption{TabArena Elo top-15 leaderboard on the full 38-dataset classification suite (200 bootstrap rounds, anchored at default RF = 1000). Light-blue/orange/green grouped bars show the Default, Tuned, and Tuned + Ensembled configurations of each method, and the highlighted yellow bar marks the TabICLv2 official default baseline. Our binary two-stage checkpoint (TabICLv2-Custom-Binary-Prior, dark orange) reaches 1590.1 Elo, $+22.1$ above the official default (1568.0).}
\label{fig:tabarena-elo}
\end{figure}

% =============================================================================
\section{Introduction}

Tabular foundation models (TFMs), namely TabPFN~\cite{hollmann2023tabpfn,hollmann2025accurate}, TabICL~\cite{qu2025tabicl,qu2026tabiclv2}, TabDPT~\cite{ma2025tabdpt}, and TabFM~\cite{kong2026tabfm}, pretrain a transformer on synthetic priors (TabDPT instead uses real data with self-supervised retrieval) so that it predicts on an unseen table in a \textbf{single forward pass with no gradient updates}~\cite{muller2022transformers}. The central mechanism is in-context learning (ICL): training rows and test rows are fed together as a \emph{context}, and the model produces the test-row labels within the same forward pass.

A question long treated as an \emph{architectural constraint} is: in classification, what determines \emph{how many classes the model can output}? In classical supervised learning the answer is immediate: the width of the output head, fixed by the number of classes in the training set, which then bounds the number of classes at inference time. The ICL paradigm breaks this constraint. TabICLv2 samples the class count of its classification prior during pretraining, typically with an upper bound of ten, so a natural assumption is that this ``ten'' is simultaneously a prior-sampling parameter and a \emph{latent capability boundary} of the model.

This paper presents a counterexample to that assumption and addresses three questions:
\begin{itemize}
\item \textbf{Q1 (phenomenon).} If the classification prior is fixed to two classes, can the fully pretrained weights still perform multi-class classification?
\item \textbf{Q2 (mechanism).} If so, what design decouples the \emph{cardinality of the output space} from the \emph{cardinality of the pretraining label space}?
\item \textbf{Q3 (cost).} What are the benefits and the failure boundaries of doing this?
\end{itemize}

Our observation is that the answer to Q1 is \textbf{yes}. On the 38 TabArena~\cite{erickson2026tabarena} classification datasets (200-bootstrap-round Elo, anchored at default RF = 1000), binary two-stage weights retrained from scratch for 540K steps (stage~1: 500K $+$ stage~2: 40K; 10K fewer than the official three-stage schedule) with bagging inference reach \textbf{1590.1 Elo} (Figure~\ref{fig:tabarena-elo}), above the official default at \textbf{1568.0} ($\boldsymbol{\Delta = +22.1}$; bootstrap intervals overlap, so the gain is directional, not significant). AUC-metric datasets are on par or slightly better; log-loss-metric datasets show a slight calibration degradation (\S\ref{sec:break}), as predicted. A production-scale validation on 60 internal credit-scoring institutions (\S\ref{sec:credit}) makes the advantage statistically significant (paired Wilcoxon $p = 0.0007$ for AUC and $p = 0.0002$ for KS).

\paragraph{Contributions.} This paper contributes:
\begin{enumerate}
\item \textbf{An unexpected empirical result.} Binary-only prior pretraining does not sacrifice multi-class inference capability (see Table~\ref{tab:leaderboard}): the binary two-stage recipe (540K steps, no stage~3) scores 1590.1 vs.\ the official default's 1568.0 Elo (550K steps over three stages). This supports treating ``ten classes'' as a \textbf{tunable hyperparameter} rather than a \textbf{capability boundary}.
\item \textbf{A mechanistic explanation.} Classification in TabICLv2 is carried by an \emph{in-context comparison operator}, with $K$ set by the labels in the context. Whether this holds \emph{depends on the label-encoder design}: a shared encoder extrapolates, a per-class embedding table does not (\S\ref{sec:hypotheses}); we give a code-level verification checklist.
\item \textbf{A cost--benefit argument.} Holding everything else fixed, a binary prior is structurally cheaper to generate and cache than a ten-class prior: it shrinks the per-task label-encoder input distribution, the label-embedding lookup, and the cached prior tensors carrying label-side state. Since prior-generation cost is cliff-shaped rather than linear (\S\ref{sec:cost}), this reduction buys a disproportionately long additional cache window, expanding the feasible frontier of long-horizon pretraining.
\item \textbf{A failure boundary.} Capping the class count too low is not free: measurable degradation appears when $K$ is large, when classes are severely imbalanced, and when class labels are ordinal. We lay out these failure scenarios and the preliminary evidence for the calibration-related ones (\S\ref{sec:break}).
\end{enumerate}

% =============================================================================
\section{Related Work}

\paragraph{Tabular Foundation Models and In-Context Optimization.}
Prior-Data Fitted Networks~\cite{muller2022transformers} pretrain transformers on synthetic datasets so that a single forward pass approximates Bayesian in-context prediction~\cite{garg2022what,li2023transformers,vonoswald2023uncovering}. TabPFN~\cite{hollmann2023tabpfn} introduced the paradigm for small classification tables; later work extended it to regression and longer contexts~\cite{hollmann2025accurate, grinsztajn2026better} and made attention linear-time via inducing points~\cite{qu2025tabicl,qu2026tabiclv2}. TabICLv2~\cite{qu2026tabiclv2} belongs to this lineage, driving both classification and regression by \emph{synthetic priors}. This paper examines a setting the lineage accepts by default but has never systematically tested: \emph{the cardinality of the prior's label space}.

\paragraph{Label Representation in ICL Predictors.}
In ICL predictors, how labels are represented determines how the model ``understands'' classes. Two dominant designs:

\begin{center}
\begin{tabularx}{\linewidth}{@{}>{\bfseries}p{0.32\linewidth} p{0.30\linewidth} >{\raggedright\arraybackslash}p{0.32\linewidth}@{}}
\toprule
Design & Mechanism & Output-space cardinality \\
\midrule
Fixed output head & A final linear map to $C$ logits, with $C$ set by configuration & Hard constraint: $C$ must be $\geq$ the number of classes at inference \\
\addlinespace
Shared label encoder + comparison decoder & Labels are mapped to embeddings by a shared encoder; predictions arise from similarity between query embeddings and each class embedding & Soft constraint: $K$ is determined by the distinct labels in the context, and extrapolates \\
\bottomrule
\end{tabularx}
\end{center}

Our core claim is that \emph{TabICLv2 belongs to the second family}, which is exactly why a binary prior still predicts multi-class. ``Class-count extrapolation'' has been glimpsed in text and vision ICL, but \emph{systematic evidence on tabular TFMs at the scale of a full pretraining run} is missing.

\paragraph{Class Count as a Prior Hyperparameter.}
A classification prior's class-count distribution affects both task diversity and generation cost. Prior work universally caps it at ten---matching the class counts of common TabArena/OpenML datasets---yet rarely asks \emph{what that cap actually constrains}. We split the question into its capability and cost halves.

\paragraph{Architecture-Side Fixes for the Class-Count Limit.}
Two recent lines attack the same ceiling from the model side. Wu and Bergman~\cite{wu2025mixture} propose a mixture-block architecture for tasks with an arbitrary number of classes, calling the cap ``the class size limitation---a crucial weakness of prior deep tabular zero-shot learners'' and patching it architecturally. Bouadi et al.~\cite{bouadi2025biaxial} pair a bi-axial attention encoder with a label-aware ICL head that scales to large label spaces via hierarchical decision routing (Orion-Bix). Both treat the cap as a tunable knob but reach that by adding machinery (a routing/mixing block, a label-aware head); ours instead reshapes the prior's class-count distribution and leaves the TabICLv2 backbone intact. The two are complementary: an architecture-side fix supplies the operators for $K$-way inference, a prior-side fix reshapes the supervision they see.

% =============================================================================
\section{Methodology}

\subsection{Preliminaries: The TabICLv2 Forward Pass}

Given $(X_{\text{train}}, y_{\text{train}}, X_{\text{test}})$, TabICLv2 executes the following pipeline (Figure~\ref{fig:pipeline}):

\begin{figure}[H]
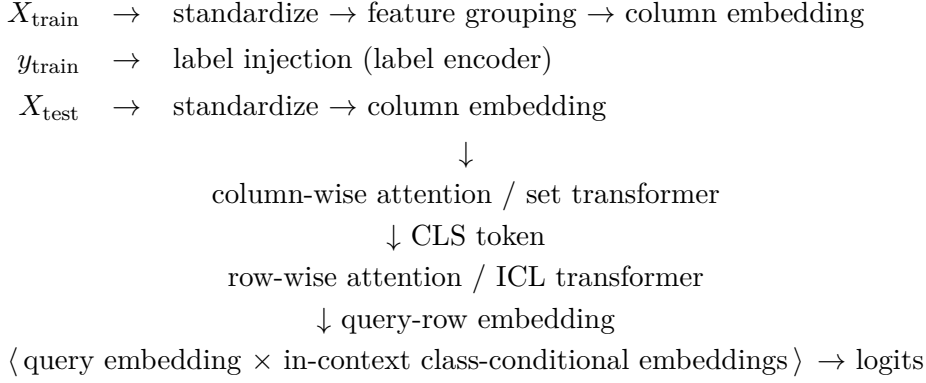

\centering
\begin{tabular}{r c l}
$X_{\text{train}}$ & $\rightarrow$ & standardize $\rightarrow$ feature grouping $\rightarrow$ column embedding \\[4pt]
$y_{\text{train}}$ & $\rightarrow$ & label injection (label encoder) \\[4pt]
$X_{\text{test}}$  & $\rightarrow$ & standardize $\rightarrow$ column embedding \\[4pt]
\multicolumn{3}{c}{$\downarrow$} \\[2pt]
\multicolumn{3}{c}{column-wise attention / set transformer} \\[2pt]
\multicolumn{3}{c}{$\downarrow$ CLS token} \\[2pt]
\multicolumn{3}{c}{row-wise attention / ICL transformer} \\[2pt]
\multicolumn{3}{c}{$\downarrow$ query-row embedding} \\[2pt]
\multicolumn{3}{c}{$\langle\,$query embedding $\times$ in-context class-conditional embeddings$\,\rangle\,\rightarrow$ logits}
\end{tabular}
\caption{TabICLv2 forward pass schematic. Standardization, feature grouping, column embedding, label injection, column-wise and row-wise attention, and the final query-to-class comparison step are shown. The two components that govern the question of this paper are the label encoder and the comparison decoder (\S\ref{sec:hypotheses}).}
\label{fig:pipeline}
\end{figure}

Two steps matter most:
\begin{itemize}
\item \textbf{Label injection.} Training-row labels are encoded and concatenated onto their rows as ``worked examples'' in the context; the \emph{form} of the encoder determines whether the class count is bound.
\item \textbf{Comparison decoding.} Logits come from comparing the query-row embedding against the class-conditional embeddings (or prototypes) in the context. Its inputs are \emph{the classes present in the context}, not \emph{the classes the configuration declared}.
\end{itemize}

Pretraining must therefore teach \emph{how to compare} (an operator), not \emph{which of ten classes to recognize} (a table).

\subsection{Where Does ``Ten Classes'' Come From?}
\label{sec:ten-classes}

The ``target class count'' appears in three distinct places in the TabICLv2 pipeline, and conflating them is the usual misunderstanding:

\begin{center}
\begin{tabularx}{\linewidth}{@{}>{\bfseries}p{0.26\linewidth} p{0.45\linewidth} >{\raggedright\arraybackslash}p{0.22\linewidth}@{}}
\toprule
Location & Role & Hard constraint? \\
\midrule
Prior sampler & Decides how many classes each synthetic task draws, typically uniform over $[2, \texttt{max\_classes}]$ & $\times$ Affects only the training-task distribution \\
\addlinespace
Label-encoder width & If a per-class embedding table, its size equals \texttt{max\_classes} & $\triangle$ \emph{Potentially a hard constraint} (the critical checkpoint) \\
\addlinespace
Evaluation protocol & The class count of the benchmark dataset itself & $\times$ Unrelated to the model \\
\bottomrule
\end{tabularx}
\end{center}

\textbf{Our change touches only the first location}: we fix the classification prior's class count to two and leave everything else (lr, optimizer, weight decay, dropout, sequence-length distribution, the regression prior's pinball configuration) untouched. That location's parameter is named \texttt{max\_classes} (shared by TabICL's training scripts and the \texttt{TabICL} class).

\begin{quote}
\textbf{Corroboration: \texttt{max\_classes} is the \emph{per-forward comparison cap}, not the \emph{total number of representable classes}.} When a dataset has more classes than \texttt{max\_classes}, the TabICL family decomposes the $K$-way task into sub-problems of size $\leq \texttt{max\_classes}$ via \emph{hierarchical classification plus mixed-radix ensembling} and recombines their outputs. The output-space cardinality is thus \emph{architecturally decoupled from \texttt{max\_classes}}, which bounds a single comparison, not the number of classes the model can express. Binary-prior training does not alter this; it merely lowers each sub-problem from 10 to 2. (Whether TabICLv2 retains this verbatim is unverified.)
\end{quote}

\subsection{Why Does It Work? Three Hypotheses}
\label{sec:hypotheses}

\textbf{H1 (operator hypothesis).} Classification is comparison; the model learns a \emph{comparison operator}. Binary tasks already supply the ``match / mismatch'' supervision to learn it, after which $K$ is just how many objects enter the comparison.

\textbf{H2 (encoder hypothesis).} The label encoder is \emph{shared across classes} (e.g., a shared scalar/vector encoding plus a class-agnostic transform), not a \emph{per-class embedding table}. This is \emph{necessary for direct $K$-way inference}: with a per-class table, out-of-table classes at inference ($K > 2$) have no embedding and extrapolation must fail in a single forward pass (it can still be rescued by the hierarchical classification + mixed-radix ensembling of \S\ref{sec:ten-classes}, which decomposes the task into sub-problems of size $\leq \texttt{max\_classes}$).

\textbf{H3 (signal hypothesis).} The discriminative signal comes mainly from \emph{column- and row-wise attention} over features; label injection only supplies an \emph{alignment} signal, so binary-only pretraining does not weaken the feature-side representation.

\begin{quote}
\textbf{Code-level verification checklist} (check each against the code before trusting \S\ref{sec:experiments}):
\begin{enumerate}
\item \textbf{Label encoder}: is it \texttt{nn.Embedding(max\_classes,\ d)} (a per-class table) or a shared encoder with a class-agnostic transform?
\item \textbf{Decoder}: do logits come from a fixed linear head, or from similarity against in-context class embeddings?
\item \textbf{Use of the class index}: is the class index consumed as a \emph{positional} encoding (extrapolates) or as a \emph{learned lookup} (does not)?
\item \textbf{Over-allocation}: if a per-class table is \emph{over-allocated} (e.g., sized for \texttt{max\_classes} = 100 but trained on 2 classes), H2 becomes ``capacity over-allocation plus sparse utilization''; the conclusion can still hold, via a different mechanism.
\end{enumerate}
\end{quote}

\subsection{Experimental Protocol}

\begin{center}
\begin{tabularx}{\linewidth}{@{}>{\bfseries}p{0.30\linewidth} >{\raggedright\arraybackslash}p{0.65\linewidth}@{}}
\toprule
Dimension & Setting \\
Target independent variable & Classification prior's target class count: official default (\texttt{max\_classes=10}, three-stage schedule, 550K steps) vs.\ ours (\texttt{max\_classes=2}, from-scratch two-stage retrain, 540K steps) \\
Current measured recipe & Ours is a from-scratch retrain of stage~1 $+$ stage~2 (500K $+$ 40K $=$ 540K) with bagging; the official default adds a 10K-step stage~3 (550K). \emph{A strictly single-variable run (class count the sole difference) is in progress} (\S\ref{sec:main}, \S\ref{sec:limitations}) \\
Controlled variables & lr, optimizer, weight decay, dropout, micro-batch, grad-accum, sequence-length distribution, recompute True (full diff in Appendix~\ref{app:config}) \\
Prior supply & Offline prior cache (target: covering all 500K steps) to avoid an order-of-magnitude latency artefact from online generation \\
Evaluation suite & TabArena classification subset, \emph{38 datasets} (binary + multiclass) \\
Metrics & Elo (Bradley--Terry, \emph{200 bootstrap rounds}, anchored to default RF = 1000), AUC / log-loss (assigned per dataset; $\pm$ std over three-fold CV) \\
Protocol & Three-fold outer CV per dataset; comparison uses \emph{final weights only} (no fine-tuning, no post-hoc calibration; bagging enabled in all TabICLv2 configurations) \\
\bottomrule
\end{tabularx}
\end{center}

% =============================================================================
\section{Experiments}
\label{sec:experiments}

\subsection{Setup}

We evaluate on \textbf{38} TabArena~\cite{erickson2026tabarena} classification datasets (binary + multiclass), strictly zero-shot (no gradient updates or post-hoc calibration; bagging enabled throughout). Elo is the Bradley--Terry rating over 200 bootstrap rounds anchored at default RF = 1000; per-dataset metrics are AUC or log-loss, each with $\pm$ std over three-fold CV.

\subsection{Main Results}
\label{sec:main}

\begin{table}[H]
\centering
\small
\caption{TabArena classification leaderboard (38 datasets, 200 bootstrap rounds, anchored at default RF = 1000). Elo+ / Elo$-$ are bootstrap intervals; Rank is the mean rank; Win rate is the pairwise win ratio. \emph{Prior $K$} is the classification prior's target class count (for TabFM, the cap reported in~\cite{kong2026tabfm}; for the other reference checkpoints, the paper's default). Steps gives total pretraining steps: the official default runs the full three-stage schedule (500K $+$ 40K $+$ 10K $=$ 550K); our Custom variant is a from-scratch retrain of stage~1 $+$ stage~2 only (540K, no stage~3). The four reference rows (TabFM, TabPFN-v2.6, the TabICLv2 official default, RealTabPFN-v2.5) use their publicly reported Elo without rerunning; ``---'' marks the original published schedule.}
\label{tab:leaderboard}
\begin{tabular}{@{}p{5.0cm} c c r r r r@{}}
\toprule
Configuration & Prior $K$ & Steps & Elo $\uparrow$ & Elo+ / Elo$-$ & Rank $\downarrow$ & Win rate \\
\midrule
TabFM (reference) & 10 & --- & 1749.6 & +133.5 / $-114.7$ & 9.59 & 0.901 \\
\textbf{TabICLv2-Custom-Binary-Prior (ours)} & \textbf{2} & 540K & \textbf{1590.1} & +75.2 / $-63.3$ & 17.91 & 0.806 \\
TabPFN-v2.6 default (reference) & 10 & --- & 1577.1 & +60.6 / $-56.2$ & 18.77 & 0.796 \\
TabICLv2 official default (baseline) & 10 & 550K & 1568.0 & +76.5 / $-66.8$ & 19.40 & 0.788 \\
RealTabPFN-v2.5 default (reference) & 10 & --- & 1521.6 & +54.7 / $-52.1$ & 22.80 & 0.749 \\
\bottomrule
\end{tabular}
\end{table}

\begin{figure}[H]
\centering
\includegraphics[width=\textwidth]{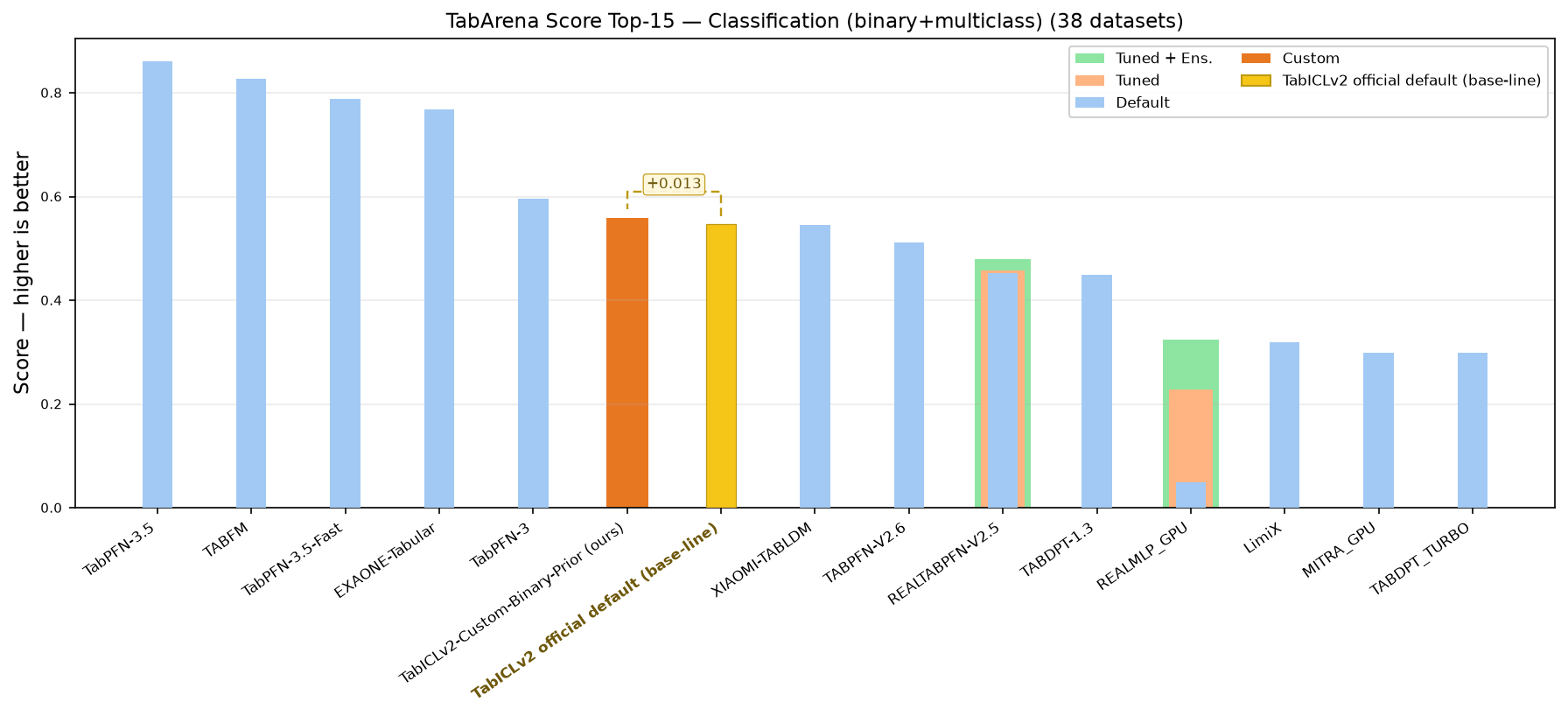}
\caption{TabArena score on the full 38-dataset classification suite (binary + multiclass; higher is better), covering all evaluated methods. Our binary two-stage checkpoint (TabICLv2-Custom-Binary-Prior, dark orange) scores $+0.013$ above the TabICLv2 official default baseline (yellow); the Default / Tuned / Tuned + Ensembled bars show the per-method configurations reported by TabArena.}
\label{fig:tabarena-score}
\end{figure}

\paragraph{Finding.}
The binary two-stage recipe scores 1590.1 vs.\ the official default's 1568.0 (a $\boldsymbol{\Delta = +22.1}$ directional gain), with rank (17.91 vs.\ 19.40) and win rate (0.806 vs.\ 0.788) moving the same way. The bootstrap intervals overlap (+75.2/$-$63.3 vs.\ +76.5/$-$66.8), so this is a \emph{directional advantage, not a significant difference}. The attribution boundary matters: ours is a from-scratch retrain of stage~1 $+$ stage~2 only (540K, no stage~3) with bagging, whereas the default adds a 10K-step stage~3, and recipe-side hyperparameters were not verified item-by-item, so the binary prior's contribution remains unisolated (Limitation~2). Even so, cutting the class count to two \emph{costs no observable Elo} against the official default---if anything $+22.1$ with 10K fewer steps---while saving prior cost (\S\ref{sec:cost}). Figure~\ref{fig:tabarena-score} puts all methods on one score axis (higher is better): ours (dark orange) directly above the official default, with a score margin of $+0.013$. The credit-scoring evaluation (\S\ref{sec:credit}) supplies the paired significance the Elo comparison lacks.

\subsection{When Does It Break?}
\label{sec:break}

\begin{center}
\begin{tabularx}{\linewidth}{@{}p{0.28\linewidth} p{0.22\linewidth} >{\raggedright\arraybackslash}p{0.45\linewidth}@{}}
\toprule
Scenario & Expected risk & Mechanism \\
\midrule
Large $K$ ($\gg$ cap) & Accuracy degradation & Crowded class embeddings in context; too few samples per class \\
Severe class imbalance & Lower minority-class recall & The comparison decoder estimates rare-class ``prototypes'' with high variance \\
Ordinal classes & Changed threshold behaviour & Binary prior only ever sees a single cut point, never multi-cut structure \\
Well-calibrated probabilities required & Log-loss degrades before accuracy & Narrower prior class distribution $\to$ mismatched prior shape on output probabilities \\
Regression tasks & No effect (expected) & Regression uses a pinball quantile loss, independent of the classification prior's class count; regression metrics were unaffected in our runs \\
\bottomrule
\end{tabularx}
\end{center}

\paragraph{Preliminary empirical evidence (binary recipe vs.\ official default).}
Row~4 is already supported by the qualitative pattern: log-loss datasets show slight calibration degradation, while AUC-scored datasets show parity or slight improvement. This \emph{separation of discrimination (unaffected) from calibration (slightly degraded)} matches the mechanism: the comparison decoder's probability \emph{shape} depends on the prior's class distribution, while the \emph{ranking} does not. Calibration-sensitive applications (e.g., credit-scoring probability outputs) need dedicated post-hoc calibration.

\subsection{Cost Analysis: Why This Matters in Practice}
\label{sec:cost}

\paragraph{Why a binary prior is cheaper in principle.}
Prior-generation cost has three structural drivers: (1) the label space (distinct classes per task); (2) the number of tasks drawn from it; and (3) the per-task cost (SCM sampling, label rendering, post-processing). Holding (2) and (3) fixed, lowering (1) from ten to two reduces the distinct label combinations the prior must support. In the TabICLv2 stack this shrinks the label-encoder input distribution, the label-embedding lookup, and the cached prior tensors carrying label-side state---all consequences of declaring $\texttt{max\_classes}=2$ at the prior sampler (\S\ref{sec:ten-classes}), with no extra engineering.

\paragraph{Why this matters in practice: the cache cliff.}
In real TabICLv2 training, prior tensors are precomputed into an offline cache; once the cache is exhausted the trainer falls back to \emph{online} generation, which costs roughly an order of magnitude more per step. In a separate (uncontrolled) observation during our 500K-step run, $\texttt{prior\_time}$ jumps from $0$ s (cache hit) to $\approx 11.5$ s (cache miss) once the cached window is exceeded, with the rest of the step at $\approx 1.4$ s. \emph{This is an observed step-time breakdown, not a controlled cap-10 vs.\ cap-2 comparison.} It does establish the engineering asymmetry: because the cliff is sharp rather than linear, any prior-cost reduction---however small---buys a \emph{disproportionately long} additional cache window. Lowering the class count from ten to two is such a structural reduction (item~1), so it lands on the steep part of the cliff; that is the engineering meaning of ``Two Is Enough.''

\subsection{Production-Scale Credit Scoring Evaluation (60 Institutions)}
\label{sec:credit}

\paragraph{Setup.}
Beyond TabArena, we evaluate two checkpoints on an internal suite of \textbf{60 real-world credit-scoring institutions} (binary default-risk scoring; 300 features each; training sizes $n \approx 1$K--$17$K; identifiers omitted for confidentiality): the \textbf{official TabICLv2 default} (prior $K=10$; full three-stage schedule) and \textbf{our binary-prior checkpoint} (from-scratch two-stage retrain, 540K steps; the same checkpoint as Table~\ref{tab:leaderboard}). Each institution uses a train/test hold-out, and we report \textbf{test-set AUC and KS}, the standard credit-risk discrimination metrics. As on TabArena the two differ in more than the class cap (ours lacks stage~3, 540K vs.\ 550K steps), so this comparison shares the attribution boundary of Limitation~2.

\begin{table}[H]
\centering
\small
\caption{Production credit-scoring suite: 60 institutions (300 features each), institution-equal-weighted macro averages on the test hold-out. $\Delta$ is ours minus the official default. Across institutions, ours wins on test AUC in 44 / ties in 1 / loses in 15 (strict sign); the paired two-sided Wilcoxon signed-rank test (normal approximation with tie-averaged ranks) gives $p = 0.0007$ on AUC and $p = 0.0002$ on KS.}
\label{tab:credit-macro}
\begin{tabular}{@{}l c c c c c@{}}
\toprule
Checkpoint & Prior $K$ & AUC$_{\text{test}}$ & KS$_{\text{test}}$ & Train$-$test gap & Cost (s) \\
\midrule
official default (baseline) & 10 & 0.7014 & 0.3082 & 0.2784 & 23.94 \\
\textbf{ours (binary, 540K)} & \textbf{2} & \textbf{0.7033} & \textbf{0.3134} & \textbf{0.2770} & 24.30 \\
\midrule
$\Delta$ (ours $-$ default) & & $+0.0019$ & $+0.0053$ & $-0.0014$ & $+0.36$ \\
\bottomrule
\end{tabular}
\end{table}

\paragraph{Findings.}
Table~\ref{tab:credit-macro} reports the macro averages and the paired statistics. Three observations:

\begin{enumerate}
\item \textbf{Significance.} Ours improves test AUC on 44 of 60 institutions ($\Delta$AUC $= +0.0019$, $\Delta$KS $= +0.0053$ macro), paired-significant at $p = 0.0007$ for AUC and $p = 0.0002$ for KS (Table~\ref{tab:credit-macro}). Unlike the directional-only TabArena Elo gap (\S\ref{sec:main}), the 60-institution paired design has the power to resolve an effect of this size.
\item \textbf{Where the gain lives.} The improvement concentrates where the baseline already performs well: terciles by baseline test AUC give $\Delta$AUC $= +0.0033$ (top, baseline $0.79$), $+0.0018$ (middle), $+0.0006$ (bottom, baseline $0.61$). Only 3 institutions show $\Delta$AUC $> +0.01$ (all mid-sized, $n = 2.7$K--$17$K), and the two largest degradations ($-0.013$, $-0.027$) both fall at the two smallest ($n = 962$, $1{,}169$; baseline AUC $\approx 0.59$--$0.61$), where single-split variance dominates. The binary prior thus \emph{improves fitting where signal exists} rather than creating capability on weak-signal institutions.
\item \textbf{Overfitting and inference cost.} The train$-$test AUC gap of ours ($0.2770$) is slightly \emph{below} the official default's ($0.2784$): cutting the prior's class count to two does not widen generalization error on this suite. Inference cost rises by $0.36$ s per institution ($+1.49\%$ over $23.94$ s), i.e., no material architectural overhead at prediction time (inference cost, distinct from the prior-generation cost of \S\ref{sec:cost}).
\end{enumerate}

% =============================================================================
\section{Discussion and Limitations}
\label{sec:limitations}

\paragraph{Why this is ``supposed'' to be true.}
If classification were genuinely carried by a $K$-dimensional output head, then ``pretraining saw only two classes'' would necessarily lock $K \leq 2$. Observing extrapolation means the causal direction is the opposite: \emph{the labels in the context define the output space, and the output head is an implementation detail.} This offers a cleaner mental model of ICL predictors: pretraining teaches \emph{how to infer a decision rule from examples}, not \emph{a fixed classifier}.

\paragraph{Limitations.}
\begin{enumerate}
\item \textbf{We establish that ``two suffices,'' not what is optimal.} A cap of 3--5 may be the more robust default.
\item \textbf{Recipe confounding.} The binary configuration (1590.1) is a from-scratch retrain of stage~1 $+$ stage~2 only (540K, no stage~3) with bagging, whereas the official default (1568.0) runs the full three-stage schedule (550K); recipe-side hyperparameters were additionally not verified item-by-item (exact optimizer and bagging settings, etc.), so training length, schedule, and recipe all differ alongside the class cap. The data support ``the binary prior shows no observable loss against the official default,'' not ``the class cap alone causes the gain.'' A strictly single-variable run (schedule- and bagging-matched, class count the sole difference) is in progress.
\item \textbf{The task-difficulty distribution is not strictly deconfounded.} Binary tasks may have a different synthetic SCM complexity distribution than ten-class tasks, so ``a weaker task mixture trained an equally good model'' cannot be excluded without controlling complexity. Two references frame the risk: Chan et al.~\cite{chan2022distributional} show that training-data distributional properties (long tail, burstiness, rare classes) \emph{drive} emergent ICL, so a binary prior also narrows the long-tail/rare-class surface the model sees; Ravent\'os et al.~\cite{raventos2023pretraining} identify a task-diversity threshold below which ICL collapses to a Bayesian estimator mimicking the pretraining distribution, so ``a smaller cap does not hurt'' requires staying above that threshold under cap-2. We control for neither.
\item \textbf{Small institutions are noise-dominated.} Institutions with $n < 2{,}000$ training rows show high single-split variance, and every large degradation in our suite falls there, so per-institution results at that scale are noise-dominated (\S\ref{sec:credit}).
\end{enumerate}

\paragraph{Future work.}
\begin{itemize}
\item \textbf{Class-count curricula}: binary first, then progressively more classes; does this beat a fixed cap of 2 or 10?
\item \textbf{Model-adaptive class count}, driven by a structural prior rather than configured.
\item \textbf{Schema-aware routes}: class semantics (e.g., an ICD hierarchy) injected by an LLM rewriting the pipeline (the TabFM-Auto route~\cite{fu2026tabfmauto}) or, conceivably, through the prior's class-count distribution.
\end{itemize}

% =============================================================================
\section{Conclusion}

We split the ``target class count cap'' of TabICLv2 pretraining---a setting accepted by default---into its capability-boundary and cost-parameter halves, and report an unexpected result: \emph{fixing the prior's target class count to two still yields weights that perform $K$-way zero-shot classification} (1590.1 vs.\ the official default's 1568.0 Elo, $\boldsymbol{\Delta = +22.1}$, a directional advantage; and a small but statistically significant test-AUC and KS gain on a 60-institution credit-scoring suite, paired Wilcoxon $p = 0.0007$ for AUC and $p = 0.0002$ for KS, at $+1.5\%$ inference cost), because classification is carried by an in-context comparison operator and the output-space cardinality is set by the context labels rather than the pretraining label space. This demotes ``number of classes'' from an architectural constraint to a tunable hyperparameter and, in principle, lowers prior generation and cache cost, expanding the feasible frontier of long-horizon pretraining. We also state the conditions under which it holds (label-encoder and decoder design), the attribution boundary of the current $\boldsymbol{+22.1}$ (ours is a from-scratch 540K two-stage retrain without stage~3, the default adds a 10K-step stage~3 on the same stage~1 $+$ stage~2, and recipe-side hyperparameters were not verified item-by-item, so the gain is not attributable to the class cap alone), and its failure boundaries (large $K$, strong imbalance, ordinal structure, and calibration, the last already with preliminary evidence).

% =============================================================================
\section*{Acknowledgments}

The authors thank Bairong Inc. for its support of this work, including the compute resources and engineering support for the pretraining runs and the production-scale evaluation.

% =============================================================================
\appendix

\section{Configuration Diff (Target Independent Variable)}
\label{app:config}

\begin{lstlisting}[language=bash,caption={Configuration diff between the official default baseline and our binary two-stage recipe. The default runs the full three-stage schedule (500K $+$ 40K $+$ 10K $=$ 550K steps); our Custom variant retrains stage~1 $+$ stage~2 only (540K, no stage~3).}]
# official default (baseline; 1568.0 Elo; 10-class prior)
--max_classes 10                  # full official schedule: stage 1 (500K)
                                  # + stage 2 (40K) + stage 3 (10K) = 550K

# ours (binary two-stage, Custom-Binary-Prior; 1590.1 Elo)
# from-scratch retrain of stage 1 (500K) + stage 2 (40K), both
# under the binary prior; no stage 3
--max_classes 2
--bagging <enabled>               # bagging inference

# everything else (--recompute True / --use_flash_attn3 False / lr /
#  optimizer / weight_decay / dropout / micro_batch / grad_accum /
#  sequence-length distribution): intended identical to the official run
\end{lstlisting}

\begin{quote}
\textbf{On the parameter name.} The target independent variable is named \texttt{max\_classes}, shared by TabICL's training scripts (\texttt{train/run.py}, \texttt{scripts/train\_stage*.sh}) and the \texttt{TabICL} model class.
\end{quote}

% =============================================================================


\begin{thebibliography}{99}

\bibitem{wu2025mixture}
Y.\ Wu and D.\ L.\ Bergman.
\newblock Zero-shot meta-learning for tabular prediction tasks with adversarially pre-trained transformer.
\newblock In \emph{ICML}, 2025.

\bibitem{bouadi2025biaxial}
M.\ Bouadi, P.\ Seth, A.\ Tanna, and V.\ K.\ Sankarapu.
\newblock Orion-Bix: Bi-axial attention for tabular in-context learning.
\newblock \emph{arXiv:2512.00181}, 2025.

\bibitem{chan2022distributional}
S.\ C.\ Y.\ Chan, A.\ Santoro, A.\ K.\ Lampinen, J.\ B.\ Wang, A.\ Singh, P.\ H.\ Richemond, J.\ McClelland, and F.\ Hill.
\newblock Data distributional properties drive emergent in-context learning in transformers.
\newblock In \emph{NeurIPS}, 2022.

\bibitem{raventos2023pretraining}
A.\ Ravent\'os, M.\ Paul, F.\ Chen, and S.\ Ganguli.
\newblock Pretraining task diversity and the emergence of non-Bayesian in-context learning for regression.
\newblock In \emph{NeurIPS}, 2023.

\bibitem{hollmann2023tabpfn}
N.\ Hollmann, S.\ Müller, K.\ Hutmacher, and F.\ Hutter.
\newblock TabPFN: A transformer that solves small tabular classification problems in a second.
\newblock In \emph{ICLR}, 2023.

\bibitem{hollmann2025accurate}
N.\ Hollmann, S.\ Müller, L.\ Purucker, A.\ Krishnakumar, M.\ Körfer, M.\ Mohr, and F.\ Hutter.
\newblock Accurate predictions on small data with a tabular foundation model.
\newblock \emph{Nature}, 2025.

\bibitem{kong2026tabfm}
W.\ Kong et al.
\newblock TabFM: A zero-shot foundation model for tabular data. Note: distinct from the separate self-evolving pipeline line TabFM-Auto by D.\ Fu et al.\ (2026, arXiv:2609.37989, see entry below).
\newblock \emph{arXiv:2609.37959}, 2026.

\bibitem{qu2025tabicl}
J.\ Qu, K.\ Holsheimer, T.\ Painter, and F.\ Hutter.
\newblock TabICL: A tabular foundation model for in-context learning on large data.
\newblock In \emph{ICML}, 2025.

\bibitem{qu2026tabiclv2}
J.\ Qu, D.\ Holzm\"uller, G.\ Varoquaux, and M.\ Le Morvan.
\newblock TabICLv2: A better, faster, scalable, and open tabular foundation model.
\newblock \emph{arXiv:2602.11139}, 2026.

\bibitem{muller2022transformers}
S.\ Müller, N.\ Hollmann, S.\ P.\ Arango, J.\ Grabocka, and F.\ Hutter.
\newblock Transformers can do Bayesian inference.
\newblock In \emph{ICLR}, 2022.

\bibitem{garg2022what}
A.\ Garg, D.\ Tsipras, P.\ Liang, and G.\ Valiant.
\newblock What can transformers learn in-context? A case study of simple function classes.
\newblock In \emph{NeurIPS}, 2022.

\bibitem{li2023transformers}
Y.\ Li, M.\ Choi, J.\ Lee, and S.\ Oh.
\newblock Transformers can do Bayesian prediction in-context.
\newblock \emph{arXiv preprint arXiv:2301.12178}, 2023.

\bibitem{vonoswald2023uncovering}
J.\ Von Oswald, M.\ Schlegel, A.\ Meulemans, S.\ Kobayashi, E.\ Niklasson, N.\ Zucchet, N.\ Scherrer, N.\ Miller, M.\ Sandler, B.\ Ag\"uera y Arcas, M.\ Vladymyrov, R.\ Pascanu, and J.\ Sacramento.
\newblock Uncovering mesa-optimization algorithms in transformers.
\newblock \emph{arXiv:2309.05858}, 2023.

\bibitem{grinsztajn2026better}
L.\ Grinsztajn, K.\ Fl\"oge, O.\ Key, F.\ Birkel, P.\ Jund, B.\ Roof, M.\ Manium, S.\ B.\ Hoo, M.\ B\"uhler, A.\ Garg, D.\ Safaric, J.\ Robertson, B.\ J\"ager, S.\ Alessi, A.\ Hayler, V.\ Moroshan, L.\ Purucker, P.\ Singer, A.\ Arazi, J.\ Siems, J.\ H.\ Metzen, G.\ Grab, N.\ Erickson, S.\ Guo, E.\ Kalfon, S.\ Bing, D.\ Salinas, C.\ Cornu, L.\ C.\ Wehrhahn, D.\ Kriuchkova, K.\ Kaya, L.\ Sidhoum, M.\ Salmon, J.\ Chen, M.\ Hulsebos, Y.\ LeCun, S.\ M\"uller, B.\ Sch\"olkopf, S.\ Gambhir, N.\ Hollmann, and F.\ Hutter.
\newblock TabPFN-3: Technical report.
\newblock \emph{arXiv:2605.13986}, 2026.

\bibitem{ma2025tabdpt}
J.\ Ma, V.\ Thomas, R.\ Hosseinzadeh, A.\ Labach, H.\ Kamkari, J.\ C.\ Cresswell, K.\ Golestan, G.\ Yu, M.\ Volkovs, and A.\ L.\ Caterini.
\newblock TabDPT: Scaling tabular foundation models on real data.
\newblock In \emph{NeurIPS}, 2025.

\bibitem{erickson2026tabarena}
N.\ Erickson et al.
\newblock TabArena: A living benchmark for machine learning on tabular data.
\newblock In \emph{NeurIPS Datasets and Benchmarks}, 2026.

\bibitem{fu2026tabfmauto}
D.\ Fu et al.
\newblock TabFM-Auto: Self-evolving pipelines for tabular foundation models.
\newblock \emph{arXiv:2609.37989}, 2026.

\end{thebibliography}
\end{document}